%% file: main.tex
\documentclass[lettersize,journal]{IEEEtran}
\usepackage{amsmath,amsfonts}
\usepackage{algorithmic}
\usepackage{algorithm}
\usepackage{color,array}
\usepackage[caption=false,font=normalsize,labelfont=sf,textfont=sf]{subfig}
\usepackage{textcomp}
\usepackage{stfloats}
\usepackage{url}
\usepackage{verbatim}
\usepackage{graphicx}
\usepackage{cite}
\usepackage{xcolor}
\usepackage[dvipsnames]{xcolor}
\usepackage{tgpagella}
\usepackage{multirow}
\usepackage{colortbl}
\usepackage[colorlinks,urlcolor=blue,linkcolor=blue,citecolor=blue]{hyperref}
\usepackage{orcidlink}
\usepackage{balance}
\usepackage{soul}
\usepackage{amssymb}
\usepackage{bm}
\usepackage{afterpage}
\usepackage{placeins}
\usepackage{todonotes}
\usepackage{makecell}
\usepackage{multirow}
\usepackage[table]{xcolor}
\usepackage{bm}  
\definecolor{customrowcolor}{RGB}{240, 240, 240}
\usepackage{stfloats}
\usepackage{pifont}

\newcommand{\xmark}{\ding{55}}
\usepackage{soul}
\usepackage[T1]{fontenc}
\usepackage{booktabs}
\usepackage{balance}

\usepackage{orcidlink}
\begin{document}

\title{MFil-Mamba: Multi-Filter Scanning for Spatial Redundancy-Aware Visual State Space Models}

\author{Puskal Khadka\,\orcidlink{0009-0005-4658-8705},  KC Santosh\,\orcidlink{0000-0003-4176-0236}
\textit{Senior, IEEE} 
        \thanks{This work was supported by the National Science Foundation under Grant No. \href{https://www.nsf.gov/awardsearch/showAward?AWD_ID=2346643}{\#2346643}, the U.S. Department of Defense under Award No. \href{https://dtic.dimensions.ai/details/grant/grant.14525543}{\#FA9550-23-1-0495}, and the U.S. Department of Education under Grant No. P116Z240151.
Any opinions, findings, conclusions or recommendations expressed in this material are those of the author(s) and do not necessarily reflect the views of the National Science Foundation, the U.S. Department of Defense, or the U.S. Department of Education.}
\thanks{Puskal Khadka and KC Santosh are with USD Artifical Intelligence Research, Department of Computer Science, University of South Dakota, Vermillion, 57069, USA. E-mail: \{puskal.khadka@coyotes.usd.edu, kc.santosh@usd.edu\} }%
}

\maketitle

\input{sec/0_abstract}    
\input{sec/1_intro}
\input{sec/2_related_work}

\input{sec/3_method}
\input{sec/4_experiment}

\input{sec/5_conclusion}

\bibliographystyle{IEEEtran}
\bibliography{main}

\begin{IEEEbiography}[{\includegraphics[width=1in,height=1.25in,clip,keepaspectratio]{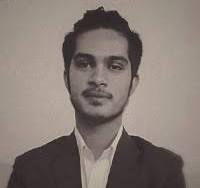}}]{Puskal Khadka} is currently an AI researcher at the USD Artificial Intelligence Research Lab. He received his M.S. in Computer Science (AI focused) from the University of South Dakota. His research interests include deep learning, computer vision, and foundational vision models, with a primary focus on self-attention mechanisms, state space models, and efficient architectures. He also works on vision-language models (VLMs) and is involved in research on methods to improve their efficiency and scalability across a wide range of visual tasks.
\end{IEEEbiography}

\begin{IEEEbiography}[{\includegraphics[width=1in,height=1.25in,clip,keepaspectratio]{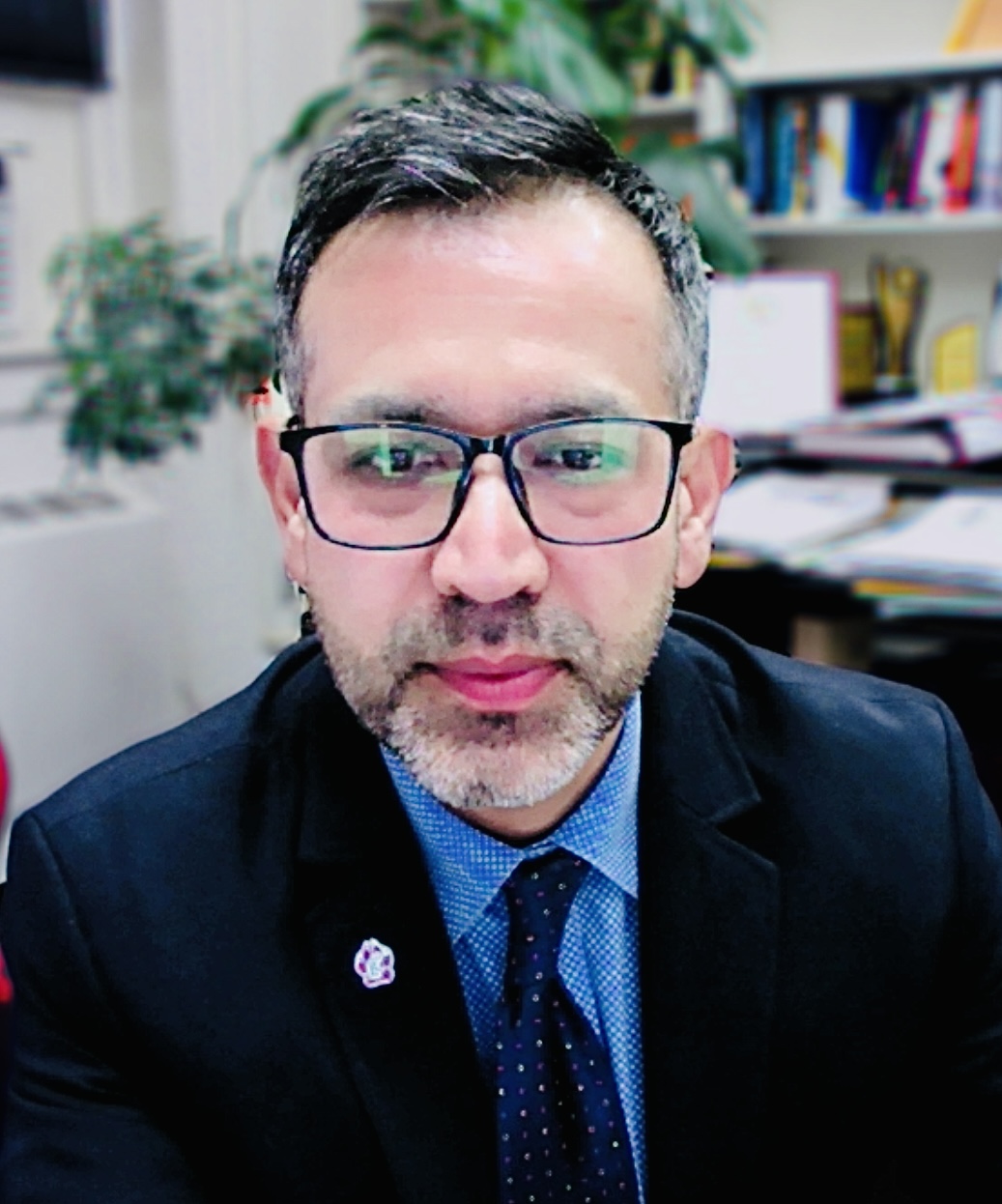}}]{KC Santosh} \textit{(Senior Member, IEEE)} – highly Accomplished AI expert – is Chair of the Department of Computer Science at the University of South Dakota (USD), where he also serves as Founding Director of the USD Artificial Intelligence Research. His extensive background includes serving as Research Fellow at NIH and Postdoctoral Research Scientist (ITESOFT co.) at LORIA Research Center, France.With over \$9 million in funding from sources such as the DOD, NSF, ED, and SDBOR, he has published 13 books and more than 300 peer-reviewed research articles, including in IEEE TPAMI, IEEE TAI, and IEEE TMI, and delivered over 100 keynote speeches. He serves as Associate Editor for prestigious journals such as IEEE Transactions on AI, IEEE Transactions on Medical Imaging, the International Journal of Machine Learning and Cybernetics, and the International Journal of Pattern Recognition \& AI. Additionally, he actively contributes to the community as Program Chair for leading conferences including the IEEE Conference on AI, CogMI, CBMS, ICDAR, and GREC, serves as a U.S. Speaker for AI education, and is a member of the NIST Center for AI Standards and Innovation.
\end{IEEEbiography}

\balance

\end{document}

%% file: sec/0_abstract.tex
\begin{abstract}
State Space Models (SSMs), especially recent Mamba architecture, have achieved remarkable success in sequence modeling tasks. However, extending SSMs to computer vision remains challenging due to the non-sequential structure of visual data and its complex 2D spatial dependencies. Although several early studies have explored adapting selective SSMs for vision applications, most approaches primarily depend on employing various traversal strategies over the same input. This introduces redundancy and distorts the intricate spatial relationships within images. To address these challenges, we propose MFil-Mamba, a novel visual state space architecture built on a multi-filter scanning backbone. Unlike fixed multi-directional traversal methods, our design enables each scan to capture unique and contextually relevant spatial information while minimizing redundancy. Furthermore, we incorporate an adaptive weighting mechanism to effectively fuse outputs from multiple scans in addition to architectural enhancements. MFil-Mamba achieves superior performance over existing state-of-the-art models across various benchmarks that include image classification, object detection, instance segmentation, and semantic segmentation. For example, our tiny variant attains 83.2\% top-1 accuracy on ImageNet-1K, 47.3\% box AP and 42.7\% mask AP on MS COCO, and 48.5\% mIoU on the ADE20K dataset. Code and models are available at \url{https://github.com/puskal-khadka/MFil-Mamba}.

\end{abstract}


%% file: sec/1_intro.tex
\section{Introduction}
\label{sec:intro}

\IEEEPARstart{V}{isual} recognition and its related downstream tasks continue to be a central area of research in deep learning. Convolutional Neural Networks (CNNs)~\cite{Krizhevsky2012alexnet,Simonyan2015vgg,he2016resnet} have long dominated image analysis by effectively learning hierarchical features representations from images. Their convolutional operations effectively capture local features and provide strong inductive biases for image analysis. However, the receptive field of convolution is inherently local, which makes it difficult for CNNs to directly model long-range dependencies within an image. The introduction of self-attention mechanism~\cite{vaswani2017attention} enabled Vision Transformers (ViTs)~\cite{dosovitskiy2020vit,liu2021swin} to capture global context between image patches and has led to enhanced performance on large-scale benchmarks. Yet, the computational cost of self-attention grows quadratically with the number of tokens, which limits its scalability and efficiency on high-resolution images and large datasets~\cite{dosovitskiy2020vit}.
\begin{figure}[tbp]
    \centering
    \includegraphics[width=1\linewidth]{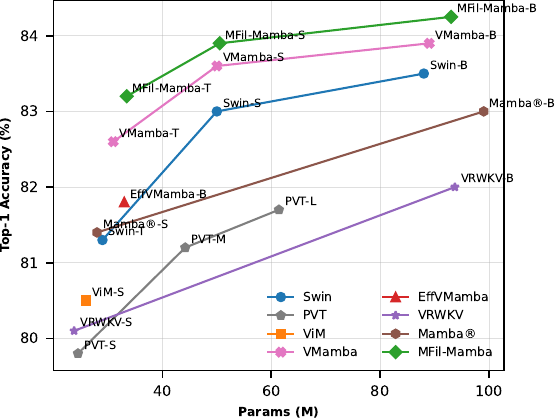}
    \caption{Top-1 Validation Accuracy versus Model Parameters comparison on Imagenet-1k~\cite{deng2009imagenet1k} datasets. MFil-Mamba demonstrates superior performance compared to baseline state-of-the-art models with similar parameter counts.}
    \label{fig:comparision}
\end{figure}

Recent models such as RWKV~\cite{peng2023rwkv} and Mamba~\cite{gu2023mamba} have emerged as efficient sequence modeling frameworks to address the computational limitations of Transformer architectures. Specifically, Mamba~\cite{gu2023mamba} achieves linear time complexity and surpasses several Transformer-based models in language and signal processing tasks. However, Mamba is originally designed for one-dimensional sequences, while visual data lie in a two-dimensional spatial domain without an inherent sequential structure, which poses challenges for applying state-space models to image data~\cite{vim,vmamba}.

To address this limitation, prior studies have introduced various directional scanning strategies to impose sequential order on image grids, such as bidirectional scanning~\cite{vim}, four-directional cross scanning~\cite{vmamba}, continuous traversal paths~\cite{yang2024plainmamba}, and more complex patterns such as Hilbert curves~\cite{Xiao2025fractablmamba, he2024mambaad}, spiral path~\cite{tang2025spiral}, and zigzag scanning~\cite{zigma}. Although partially effective, these approaches often introduce information redundancy, spatial distortion, and poor local–global coherence.

In this work, we present MFil-Mamba, a novel architecture for visual modeling that replaces directional traversal with a multi-filter strategy. Instead of relying on predefined directional scanning paths, the proposed approach applies a set of dynamic spatial filters to extract complementary structural and contextual cues from input feature maps. The resulting filtered representations are then processed through a state-space model to capture diverse spatial dependencies without imposing an explicit traversal order. To enhance this design, we also introduce a lightweight adaptive fusion mechanism that dynamically aggregates the outputs of the different SSM-processed filtered scans. This module learns to emphasize the most informative spatial features while suppressing less relevant representations. We further explore multiple architectural variants and model configurations, demonstrating that MFil-Mamba serves as a versatile and general-purpose vision backbone for a wide range of image analysis tasks. Extensive experiments show that MFil-Mamba achieves state-of-the-art performance on ImageNet-1K~\cite{deng2009imagenet1k}, outperforming existing competitive models with similar size and computational complexity, as shown in Fig.~\ref{fig:comparision}. The key contributions are summarized as follows:
\begin{enumerate}
\item We introduce MFil-Mamba, a novel multi-filter scanning architecture that overcomes the limitations of directional traversal strategy in existing visual state-space models.
\item We incorporate a lightweight adaptive fusion mechanism along with architectural refinements to enhance spatial feature integration and improve the quality of visual representations.
\item We develop three model variants and demonstrate state-of-the-art performance across multiple vision tasks, including image classification, object detection, and segmentation benchmarks. Our detailed experimental results can be found in Sections~\ref{subsec:classification}–~\ref{subsec:semantic-seg}.
\end{enumerate}

The remainder of this paper is organized as follows. Section~\ref{sec:related_works} discusses related works in convolutional neural networks, vision transformers, and state space models. Our methodology is detailed in Section~\ref{sec:method}, and it includes preliminaries, Mfil-Mamba architecture, multi-filter scanning strategy, and theoretical justification. Experimental results, along with analysis, visualization, and ablation studies, are presented in Section~\ref{sec:exp}.

%% file: sec/2_related_work.tex
\section{Related Works}
\label{sec:related_works}
\subsection{Convolutional Neural Networks (CNNs) }
CNNs~\cite{lecun1998cnn,Krizhevsky2012alexnet,Simonyan2015vgg, szegedy2015going, he2016resnet,li2019selective,tan2019efficientnet} have long served as the cornerstone of computer vision, effectively capturing local spatial dependencies through learnable convolutional filters. Architectures such as AlexNet~\cite{Krizhevsky2012alexnet}, VGG~\cite{Simonyan2015vgg}, ResNet~\cite{he2016resnet}, and DenseNet~\cite{huang2017densely} achieved state-of-the-art performance on large-scale benchmarks such as ImageNet~\cite{deng2009imagenet1k}. Several advancements have been proposed in CNNs, such as dilated convolution~\cite{yu2016multi}, depth-wise convolution~\cite{howard2017mobilenets}, and deformable convolution~\cite{dai2017deformable}. Recent CNN architectures have explored new design space to enhance performance and efficiency. Specifically, RegNetY~\cite{radosavovic2020regnety}  investigates the network design space to identify optimal configurations, and ConvNeXt~\cite{liu2022convnet} modernizes CNNs by integrating architectural ideas inspired by vision transformers. Similarly, SLAK~\cite{liu2022more} proposes distillation of knowledge from a large-kernel ConvNet into a small-kernel network to achieve improved performance and robustness. Moreover, CNNs have been widely adopted as general-purpose visual feature extractors for downstream tasks including object detection and segmentation and form the backbone of prominent frameworks such as Faster R-CNN~\cite{ren2015fasterrcnn}, YOLO~\cite{redmon2016yolo} and Mask R-CNN~\cite{he2017maskrcnn}. 

Despite their success, the inherently local receptive fields of CNNs limit their ability to capture global dependencies. Increasing depth or kernel size can partially address this, but it often leads to higher computational complexity~\cite{lau2024large}. To overcome these limitations, hybrid approaches combining convolution with attention mechanisms and multi-scale feature extraction techniques have been developed, which allow networks to capture both fine-grained details and long-range context~\cite{wu2021cvt, dai2021coatnet}. Nevertheless, conventional CNNs may still struggle with very long-range dependencies or complex scenes, which has motivated the development of transformer-based and dynamic convolutional architectures that explicitly model global interactions.

\subsection{Vision Transformer (ViTs)}
The success of the Transformer~\cite{vaswani2017attention} architecture in natural language processing has inspired its adaptation to computer vision, leading to models such as ViT~\cite{dosovitskiy2020vit} and its follow-up variants~\cite{liu2021swin, wang2022pvt2, touvron2021deit, chu2021twins, bao2022beit, yuan2021tokens}. Many of these models have demonstrated superior performance over traditional CNNs across a range of vision benchmarks. Specifically, Swin Transformer~\cite{liu2021swin} and PVT~\cite{wang2021pyramid} propose hierarchical transformer architectures with gradually decreasing feature map sizes, while DeiT~\cite{touvron2021deit} introduces distillation-based training, where a student model learns from a teacher model through the attention mechanism. ViT-Adapter~\cite{chen2022vitadapter} introduces a pre-training-free adapter to adapt the plain Vision Transformer for multiple dense prediction tasks. DAT~\cite{xia2022vision} proposes a deformable self-attention mechanism to model sparse attention in a data-dependent manner. The key advantage of these transformer-based vision models lies in their self-attention mechanism, which effectively captures long-range dependencies by enabling global interactions between image patches. However, ViTs suffer from high computational and memory costs due to the quadratic complexity of self-attention with respect to input size, which poses scalability challenges for high-resolution visual tasks~\cite{dosovitskiy2020vit, khan2022transformers, zhang2021rest}.

Recent studies have focused on improving the efficiency and flexibility of vision transformers through local attention, patch merging, and hybrid CNN-transformer designs. Several techniques, such as data-efficient training~\cite{touvron2021deit} and parameter efficient fine-tuning~\cite{he2023parameter} have been proposed to mitigate the high computational resource requirements. Additionally, transformer models have shown promising transfer learning capabilities in multiple visual tasks such as object detection, semantic segmentation, and medical image analysis~\cite{li2021benchmarking,aburass2025vision}. Despite these advancements, balancing computational efficiency with model expressiveness remains a critical research challenge in applying transformer architectures to large-scale, real-world vision problems.

\subsection{State Space Models (SSMs) }
SSMs~\cite{gu2022efficiently,smith2022s5,gu2023mamba} have recently emerged as an efficient alternative to Transformers, offering linear scalability with sequence length. Structured state space model (S4)~\cite{gu2022efficiently} enhances SSMs by decomposing the state space matrices into the sum of low-rank and skew-symmetric components, while H3~\cite{fu2022hungry} proposes a new SSM layer based on shift and diagonal matrices to bridge the gap between SSMs and attention mechanisms for language modeling tasks. Furthermore, S5~\cite{smith2022s5} introduces parallel scanning to handle time-varying dynamics efficiently. Building on these, Mamba~\cite{gu2023mamba} incorporates a selection mechanism into the structured state space along with a hardware-aware efficient parallel algorithm. 

Due to Mamba’s linear-time efficiency~\cite{gu2023mamba}, several recent studies have explored its adaptation to computer vision~\cite{vim,vmamba,patro2024simba,xiao2024grootvl,wan2025sigma,li2024mamba,Liu2025mamba4d}. Specifically, Vision Mamba~\cite{vim} employs bidirectional scanning to process 2D image data, while VMamba~\cite{vmamba} extends this design with a four-way cross-scan scheme. PlainMamba~\cite{yang2024plainmamba} adopts a direction-aware continuous scanning strategy, and FractalMamba~\cite{Xiao2025fractablmamba} uses a fractal scanning path such as Hilbert curve to transform spatial feature maps into 1D sequences. Similarly, MSVMamba~\cite{shi2024multi} applies a multi-scale 2D scanning technique using bidirectional traversal on different scale layer, whereas EAMamba~\cite{lin2025eamamba} implements an all-around directional scanning mechanism integrating both cross and diagonal scans. Furthermore, these models have been applied to multiple vision tasks such as video analysis~\cite{li2024videomamba}, medical image segmentation~\cite{ma2024u,xing2025segmamba}, and others~\cite{shi2025vmambair,li2025alignmamba}.

To date, most approaches linearize 2D images into 1D sequences via multiple directional traversals to align with Mamba’s inherently ordered sequential processing capabilities~\cite{vim,vmamba}. The underlying assumption is that a collection of directional scans transforms 2D image data into sequentially ordered tokens enriched with contextual cues from multiple orientations, thereby making them compatible with Mamba’s inherent sequential processing architecture. However, such fixed direction-based traversal schemes often introduce redundancy and distort the original spatial dependencies.

\begin{figure*}[!t]
    \centering
    \includegraphics[width=\linewidth]{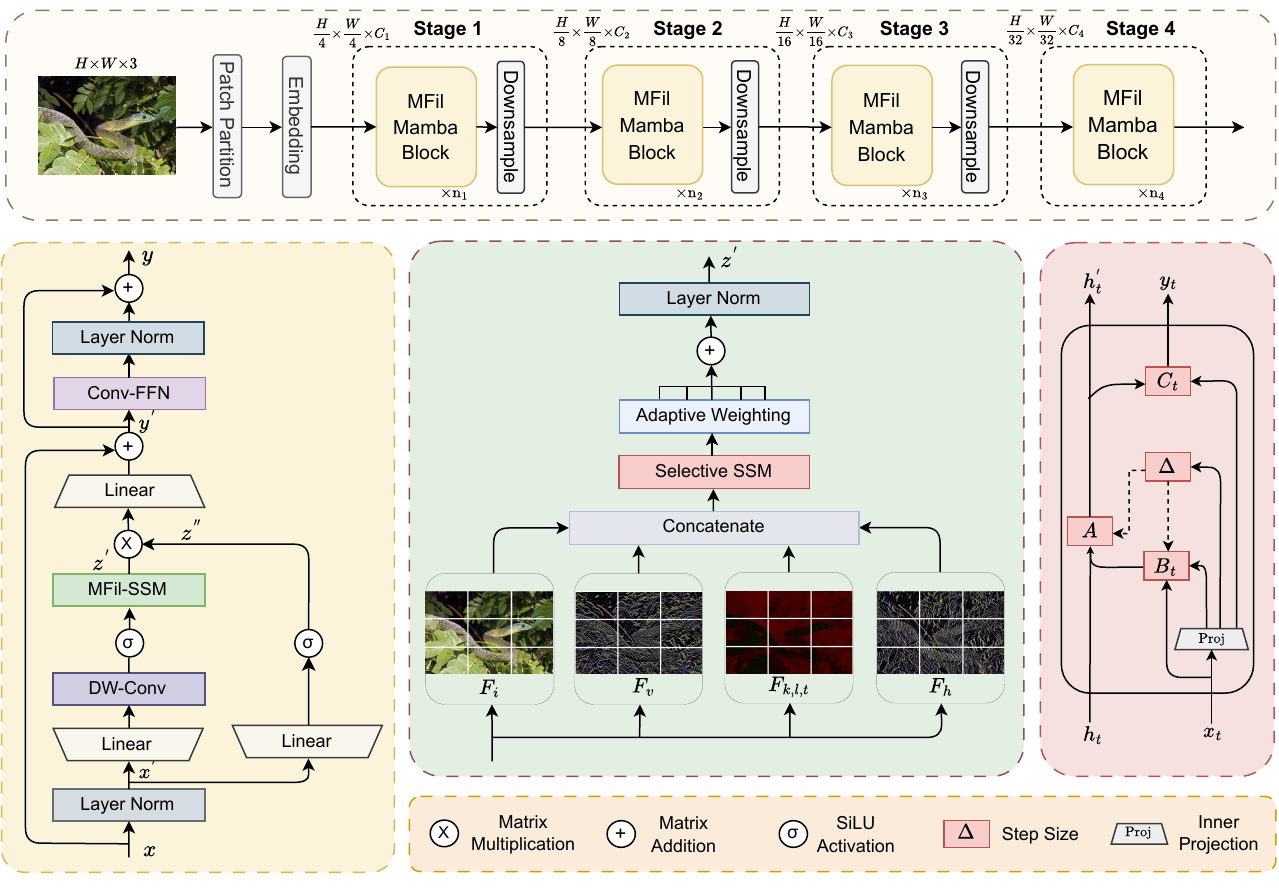}
    \caption{ \textbf{(Top)} Overview of the MFil-Mamba. \textbf{(Bottom Left)} Illustration of Single MFil-Mamba Block. \textbf{(Bottom Middle)} Illustration of MFil-SSM block with filter-based scanning across
four input representations. Each representation is independently
filtered and then its patches are concatenated and passed through a selective state space block. \textbf{(Bottom Right)} Illustration of State space model.}
    \label{fig:architecture}
\end{figure*}

In contrast, our method diverges from this direction-based scan paradigm. Rather than relying on these pre-defined directional sequential reordering, we propose a multi-filter mechanism embedded directly into the Mamba block. Each filter models spatial dependencies through distinct receptive field, which enables parallel extraction of complementary spatial features. This spatially aware multi-filter design allows Mamba to process visual data natively in the 2D domain, offering a more efficient and structurally coherent framework for a wide range of vision tasks.

%% file: sec/3_method.tex
\section{MFil-Mamba and its Variants}
\label{sec:method}

\subsection{Preliminaries}
SSM transforms a 1D continuous input sequence \(x(t) \in \mathbb{R}\) into an output sequence \(y(t) \in \mathbb{R}\) by using a hidden state \(h(t) \in \mathbb{R}^{N}\) that evolves over time. Using continuous-time ordinary differential equation, this relationship can be expressed as
\begin{align} \nonumber
    &h'(t) = \bm{A}h(t) + \bm{B} x(t), \\
    &y(t) = \bm{C}h(t),
\label{eq:ode}
\end{align}
where \(\bm{A} \in \mathbb{R}^{N \times N}\) is an evolution matrix that defines the dynamics of the hidden state \(h(t)\), \(\bm{B} \in \mathbb{R}^{N \times 1}\) maps the input \(x(t)\) into the state space, and \(\bm{C} \in \mathbb{R}^{1 \times N}\) maps the hidden state to the output \(y(t)\). These matrices \((\bm{A}, \bm{B}, \bm{C})\) are learnable parameters. To use continuous-time equation in deep learning, it is necessary to convert it into a discretized version. Using the zero-order hold (ZOH) technique with a scalar time scale parameter $\Delta$, Eq.~\ref{eq:ode} can be discretized as follow:
\begin{align} \nonumber
&h'(t) = \bm{\overline{A}}h(t-1) + \bm{\overline{B}} x(t), \\
 &y(t) = \bm{\overline{C}}h(t),
\label{eq:discretized}
\end{align}
where $\bm{\overline{A}}, \bm{\overline{B}}$ are the discrete form of continuous parameters $\bm{A}, \bm{B}$ derived using $I$ identity matrix:
\begin{align} \nonumber
    &\bm{\overline{A}} = \exp(\Delta \bm{A}), \\
    &\bm{\overline{B}} = (\Delta \bm{A})^{-1} \big(\exp(\Delta \bm{A}) - \bm{I}\big) \Delta \bm{B}.
\end{align}
As Eq.~\ref{eq:discretized} is expressed in linear recurrence form, for efficient computation and parallel training, it can be implemented through a global convolution operation, as mathematically expressed below:
\begin{align}\nonumber
        &\bm{\overline{K}} = (\bm{C\overline{B}, C \overline{A}\overline{B},\ldots,C\overline{A}^{L-1}\overline{B}}), \\
        &y = \bm{x} * \bm{\overline{K}},
    \label{eq:ssm-conv}
\end{align}
where $L$ is the input sequence length, $*$ is the convolution operation, and $\bm{\overline{K}} \in \mathbb{R}^{L}$ represents the SSM convolution kernel. For all cases, $\bm{\overline{A},\overline{B},\Delta}$ are always constant behaving as a linear time invariant (LTI) system. This LTI hinders the model's ability to focus on or ignore specific inputs~\cite{gu2023mamba}. To overcome this limitation, Mamba introduces selective scanning mechanism by using input-dependent SSM parameters along with a hardware aware algorithm over Eq.~\ref{eq:ssm-conv}.

\subsection{Model Architecture}
Our model employs a four-stage hierarchical architecture, where each stage produces uniformly down-sampled feature maps. For any input image \( I \in \mathbb{R}^{H \times W \times 3} \), it is first divided into non-overlapping \( 4 \times 4 \) patches, yielding an embedding of size \( \frac{H}{4} \times \frac{W}{4} \times C \). The feature maps then pass through \( n \) MFil-Mamba blocks per stage. Except for the final stage, a downsampling layer is added which halves the spatial resolution while simultaneously increasing the channel dimension according to the model variant, as specified in Sec.~\ref{subsec:model_variants}. To better capture spatial visual cues, we replace the conventional multilayer perceptron layer with a Convolutional Feed-Forward Network (ConvFFN), inspired by~\cite{wang2022pvt2}. The output from the final stage is passed to a task-specific prediction head.

\subsection{Single MFil-Mamba Block}
A single MFil-Mamba block consists of a filter-based selective state space module (MFil-SSM) followed by a convolution-based feed-forward network, as illustrated in the bottom-left of Fig.~\ref{fig:architecture}. Given an input feature map $x \in \mathbb{R}^{H \times W \times C}$, the block processes it through a sequence of transformations, which can be mathematically formulated as follows:
\begin{align}
    &x^{'} = \text{LN}(x), \nonumber \\
    &z^{'} = \text{MFil-SSM}\!\left(\sigma\!\left(\text{Conv}\!\left(\text{Linear}(x^{'})[\ldots, :C]\right)\right)\right), \nonumber \\
    &z^{''} = \sigma\!\left(\text{Linear}(x^{'})[\ldots, C:]\right), \nonumber \\
    &y^{'} = \text{Linear}(z^{'} \odot z^{''}) + x, \nonumber \\
    &y = y^{'} + \text{LN}\!\left(\text{ConvFFN}(y^{'})\right),
\end{align}
where \( LN \) represents layer normalization, \( C \) denotes the embedding dimension of each linear projection branch, and \( \sigma \) denotes SiLU activation function. The term \( x^{'} \) represents the output of the Layer normalization, \( z^{'} \) and \( z^{''} \) represent the outputs of the two projection branches, while \( y^{'} \) and \( y \) correspond to the outputs of the MFil-SSM and ConvFFN modules, respectively. The MFil-SSM refers to the filter-based SSM module, detailed in Section~\ref{subsec:mfilssm} (follow graphical illustrations in the bottom-left and bottom-middle of Fig.~\ref{fig:architecture}).

\subsection{Multi Filter Scanning Strategy}
\label{subsec:mfilssm}
We introduce a multi-filter scanning strategy to capture spatial relationships from diverse perspectives without relying on redundant multi-directional traversal. It improves feature representation by applying multiple filters to the input image. For any input image \( I \in \mathbb{R}^{H \times W \times C} \), three auxiliary feature maps are produced using both fixed and learnable filters. Specifically, horizontal and vertical filters are employed to form an orthogonal basis for edge information in the first-order gradient space, while an additional dynamic filter, optimized during training, enables adaptation to task-specific patterns.

\begin{enumerate}
\item Orthogonal feature maps: Horizontal \( F_{h} \) and vertical \( F_{v} \) feature maps are computed using Sobel-based kernels \( \{K_x, K_y\} \in \mathbb{R}^{3 \times 3}\), where each kernel captures gradient variations along its respective axis, i.e, 
\( K_x \) for vertical edge and \( K_y \) for horizontal edge features. The resulting filtered maps are refined using lightweight depthwise convolution to enhance structural visual cues:
\begin{equation}\nonumber
     F_{h} = \text{DWConv}(I*K_{y}), \\
     F_{v} = \text{DWConv}(I*K_{x}).
\end{equation}
 
\item Dynamic feature map: A dynamic feature map \( F_{k,l,t} \) is generated using a depthwise separable convolution block to capture high-level discriminative patterns. 
Let \(\mathbf{K}^{(d)}_{i,j,s}\) denote the spatial kernel applied to the \(s^{th}\) input channel in the depthwise stage, and let \(\mathbf{K}^{(p)}_{1,1,s,t}\) denote the \(1\times1\) kernel mapping input channel \(s\) to output channel \(t\) in the pointwise convolution stage, represented as:
\begin{align}\nonumber
    & F^{(d)}_{k,l,s} = \sum_{i,j} \mathbf{K}^{(d)}_{i,j,s} \cdot \mathbf{F}_{k+i-1,\,l+j-1,\,s}, \nonumber \\
    &F_{k,l,t} = \sum_{s=1}^{C_{\text{in}}}\mathbf{K}^{(p)}_{1,1,s,t} \cdot \mathbf{F}^{(d)}_{k,l,s}.
\end{align}
\end{enumerate}

The multi-filter gradient feature maps, together with the original input \( I \), form a four-fold representation set \( \{F_i, F_h, F_v, F_{k,l,t}\} \), where \( F_i = I \). Incorporating \( I \) preserves the original visual cues alongside the extracted gradient-based features. The combined representation is obtained by concatenating four feature maps along the feature axis, $\mathrm{q}$, expressed as:
\begin{equation}
    F_{\text{concat}} = \text{Stack$_{q}$}(F_i, F_h, F_v, F_{k,l,t}) \in \mathbb{R}^{(H \times W \times 4)\times C}.
\end{equation}
The concatenated feature map \( F_{\text{concat}} \) is subsequently fed into the selective state-space block as shown in the bottom right of Fig.~\ref{fig:architecture}.

\subsection{Adaptive Merging}
\label{subsec:adaptive_merging}
To integrate the resulting outputs from the selective state-space block, we employ an adaptive weighting mechanism with trainable normalized weights \( \alpha_i \) computed via softmax: \( \alpha_i = \frac{e^{w_i}}{\sum_{j=0}^{3} e^{w_j}}, i \in \{0,1,2,3\}\), where \( w_i \) denotes the learnable scalar for each output scan $F_i'$. The fused feature map can be expressed as \( F_{\text{fused}} = \sum_{i=0}^{3} \alpha_i F_i'\).
It enables the model to adaptively assign importance to each scan and thus enhance its ability to capture rich spatial features. The adaptive weighting further ensures that the fused representation lies within the convex hull of the transformed features, which allows the network to interpolate between different perspectives in a data-driven manner.

\subsection{Theoretical Discussion on Multi Filter Scanning}
\label{subsec:theory}
We briefly justify theoretically how our method improves representational capacity. Let \( X \in \mathbb{R}^{H \times W \times C} \) denotes a feature map, and let \( \mathbf{x} \) be its vectorized form. The spatial covariance matrix of \( \mathbf{x} \) is
\begin{equation}
\Sigma_x = \mathrm{E}\big[(\mathbf{x} - \mu_x)(\mathbf{x} - \mu_x)^{\top}\big],
\label{eq:covar1}
\end{equation}
 $\text{where } \mu_x = \mathrm{E}[\mathbf{x}].$ SSM-based vision models often serialize the feature map via directional traversal to form 1D sequences. Let \(\mathbf{x}_i\) and \(\mathbf{x}_j\) be such reordered sequences derived from original vector \(\mathbf{x}\) using permutation matrices \( P_i \) and \( P_j \): \(\mathbf{x}_i = P_i \mathbf{x}, \; \mathbf{x}_j = P_j \mathbf{x}\), and the cross-covariance matrix between them is mathematically expressed as:
\begin{equation}
\Sigma_{ij} = \mathrm{E}\big[(\mathbf{x}_i - \mu_i)(\mathbf{x}_j - \mu_j)^{\top}\big],
\label{eq:covar2}
\end{equation}
where \( \mu_i = \mathrm{E}[\mathbf{x}_i] = \mathrm{E}[P_i \mathbf{x}] = P_i \mu_x \).  
Substituting \(\mu\) and permutation relationships into Eq.~\ref{eq:covar2}, we obtain
\begin{align}
\small
\Sigma_{ij} &= \mathrm{E}\big[(P_i \mathbf{x} - P_i \mu_x)(P_j \mathbf{x} - P_j \mu_x)^{\top}\big], \notag \\ 
\Sigma_{ij} &= P_i \, \mathrm{E}\big[(\mathbf{x} - \mu_x)(\mathbf{x} - \mu_x)^{\top}\big] \, P_j^{\top} \notag.
\end{align}
Using Eq.~\ref{eq:covar1}, this simplifies to
\begin{equation}
\small
\Sigma_{ij} = P_i \, \Sigma_x \, P_j^{\top}.
\label{eq:theory_1}
\end{equation}
This result shows that directional traversal just reorders the existing spatial covariance structure. Hence, it introduces redundant representations and cannot capture new second-order dependencies, while potentially distorting spatial locality through global reordering.

To overcome these limitations, we introduce a set of learnable spatial filters $\{F\}$, each designed to extract distinct structural cues directly from the feature map. Unlike permutation operators, each filter projects the input into a distinct subspace of spatial features. For a given filter $F_i$, applied to the vectorized input feature map $\mathbf{x}$, the filtered representation will be
\(
\mathbf{y}_i = F_i * \mathbf{x},
\) where $*$ denotes convolution. Under standard vectorization, convolution can be expressed as multiplication by a structured linear operator. Therefore, the operation can equivalently be written as
\(
\mathbf{y}_i = F_i \mathbf{x}
\). Similarly, the mean of the filtered output \(\mathbf{y}_i\) will be
\(
\mu_i = \mathrm{E}[\mathbf{y}_i] = \mathrm{E}[F_i \mathbf{x}] = F_i \, \mu_x.
\)
Using the definition of covariance, the covariance of a single filtered representation is
\begin{equation}
\Sigma_i 
= \mathrm{E}\big[(F_i * \mathbf{x} - \mu_i)(F_i * \mathbf{x} - \mu_i)^{\top}\big].
\end{equation}
Extending this to two filtered representations obtained using filters $F_i$ and $F_j$, the cross-covariance between them is defined as:
\begin{align}
\Sigma_{ij}
&= \mathrm{E}\big[(\mathbf{y}_i - \mu_i)(\mathbf{y}_j - \mu_j)^{\top}\big] \notag \\
&= \mathrm{E}\big[(F_i  \mathbf{x} - F_i  \mu_x)(F_j  \mathbf{x} - F_j  \mu_x)^{\top}\big] \notag \\
&= F_i \, \mathrm{E}\big[(\mathbf{x} - \mu_x)(\mathbf{x} - \mu_x)^{\top}\big] \, F_j^{\top} \notag \\
&= F_i \, \Sigma_x \, F_j^{\top}.
\end{align}

\begin{table*}[!t]
\centering
\small
\caption{Detailed architecture overview of the \textbf{MFil-Mamba} variants. }
\begin{tabular}{l|c
|c|c|c}
\toprule[1pt]
Layer & \begin{tabular}[c]{@{}c@{}} Output Size \end{tabular} & MFil-Mamba-T  & MFil-Mamba-S & MFil-Mamba-B  \\

\hline
Stem &  56 $\times$ 56 & dim 94 & dim 94 & dim 128  \\
\hline
\hline

\multirow{3}{*}{Stage 1}  & \multirow{3}{*}{ 28 $\times$ 28} & $\begin{bmatrix}\text{MFil-Mamba Block,} \\ \text{dim 94}\end{bmatrix}$ $\times$ 1  & $\begin{bmatrix}\text{MFil-Mamba Block,} \\ \text{dim 94}\end{bmatrix}$ $\times$ 2 & $\begin{bmatrix}\text{MFil-Mamba Block,} \\ \text{dim 128}\end{bmatrix}$ $\times$ 2 \\
\cline{3-5}
&  &\makecell{Downsample, \\ conv 2 $\times$ 2, stride 2, dim 188 } & \makecell{Downsample, \\ conv 2 $\times$ 2, stride 2, dim 188 } &  \makecell{Downsample, \\ conv 2 $\times$ 2, stride 2, dim 256 } \\
\hline
\hline
\multirow{3}{*}{Stage 2}  & \multirow{3}{*}{ 14 $\times$ 14} & $\begin{bmatrix}\text{MFil-Mamba Block,} \\ \text{dim 188}\end{bmatrix}$ $\times$ 3  & $\begin{bmatrix}\text{MFil-Mamba Block,} \\ \text{dim 188}\end{bmatrix}$ $\times$ 2 & $\begin{bmatrix}\text{MFil-Mamba Block,} \\ \text{dim 256}\end{bmatrix}$ $\times$ 2 \\
\cline{3-5}
&  &\makecell{Downsample, \\ conv 2 $\times$ 2, stride 2, dim 376 } & \makecell{Downsample, \\ conv 2 $\times$ 2, stride 2, dim 376 } &  \makecell{Downsample, \\ conv 2 $\times$ 2, stride 2, dim 512 } \\
\hline
\hline

\multirow{3}{*}{Stage 3}  & \multirow{3}{*}{ 7 $\times$ 7} & $\begin{bmatrix}\text{MFil-Mamba Block,} \\ \text{dim 376}\end{bmatrix}$ $\times$ 8  & $\begin{bmatrix}\text{MFil-Mamba Block,} \\ \text{dim 376}\end{bmatrix}$ $\times$ 18 & $\begin{bmatrix}\text{MFil-Mamba Block,} \\ \text{dim 512}\end{bmatrix}$ $\times$ 18 \\
\cline{3-5}
&  &\makecell{Downsample, \\ conv 2 $\times$ 2, stride 2, dim 752 } & \makecell{Downsample, \\ conv 2 $\times$ 2, stride 2, dim 752 } &  \makecell{Downsample, \\ conv 2 $\times$ 2, stride 2, dim 1024 } \\
\hline
\hline

\multirow{1}{*}{Stage 4}  & \multirow{1}{*}{ 7 $\times$ 7} & $\begin{bmatrix}\text{MFil-Mamba Block,} \\ \text{dim 752}\end{bmatrix}$ $\times$ 2  & $\begin{bmatrix}\text{MFil-Mamba Block,} \\ \text{dim 752}\end{bmatrix}$ $\times$ 2 & $\begin{bmatrix}\text{MFil-Mamba Block,} \\ \text{dim 1024}\end{bmatrix}$ $\times$ 2 \\

\hline
\hline
Output Head &  1 $\times$ 1 & \multicolumn{3}{c}{Average Pool, dim 1000, softmax} \\
\bottomrule[1pt]
\end{tabular}
\normalsize
\label{table:model_variant}
\end{table*}

\noindent This shows that multi-filter scanning reweights and reorients the underlying covariance \(\Sigma_x\), projecting inputs into orthogonal and adaptive subspaces. This captures distinct spatial dependencies, preserves local coherence, avoids possible redundancies, and thereby underlies the theoretical basis for the improved accuracy ({\em ref.} Table~\ref{table:classification}) and effective receptive field as visualized in Section~\ref{subsec:erf}.

\subsection{Model Variants}
\label{subsec:model_variants}
We develop three variants of the MFil-Mamba architecture, i.e, Tiny (T), Small (S), and Base (B). MFil-Mamba-T is our lightest variant with 33.5M parameters, followed by the small variant, MFil-Mamba-S, having 50.5M parameters. MFil-Mamba-B is our largest variant with 93.1M parameters. Each variant has four-stage hierarchical structure. The variants differ in terms of the number of blocks per stage and the corresponding embedding dimensions. The downsampling layer at each stage is implemented using stride convolution with a $2 \times 2$ kernel and stride $2$, which reduces the spatial resolution of the feature maps by half. The detailed architectural configurations of each variant are presented in Table~\ref{table:model_variant} and the model parameters and GFLOPs are further reported in Table~\ref{table:classification}.
\begin{itemize}
    \item MFil-Mamba-T: $C = 94$, \; Layers $= \{1, 3, 8, 2\},$
    \item MFil-Mamba-S: $C = 94$, \; Layers $= \{2, 2, 18, 2\},$ and
    \item MFil-Mamba-B: $C = 128$, \; Layers $= \{2, 2, 18, 2\}.$ 
\end{itemize}
Here, $C$ denotes the channel dimension, and `Layers' indicate the number of blocks per stage.

%% file: sec/4_experiment.tex
\section{Experiments}
\label{sec:exp}
We evaluate our model on key computer vision tasks: image classification, object detection, instance segmentation, and semantic segmentation. Additionally, we provide Grad-CAM for model interpretability, ERF for receptive field analysis, and qualitative visualizations for further model analysis in real world datasets.
\begin{table}[tbp]
    \centering
    \small
      \caption{Hyperparameters of MFil-Mamba models for ImageNet-1k classification task.}
    \begin{tabular}{l|ccc}
    \toprule[1pt]
    Hyperparameter & Tiny & Small & Base \\\midrule
    Input Size & \multicolumn{3}{c}{224} \\
    Epochs & \multicolumn{3}{c}{300} \\
    Batch Size & \multicolumn{3}{c}{1024} \\
    Optimizer & \multicolumn{3}{c}{AdamW} \\
    Weight Decay & \multicolumn{3}{c}{0.05} \\
    Learning Rate Decay & \multicolumn{3}{c}{Cosine} \\
    Stochastic Drop Path & 0.2 & 0.3 & 0.5 \\
    Rand Erase Probability & \multicolumn{3}{c}{0.25} \\
    Rand Augment & \multicolumn{3}{c}{m9-msd0.5} \\
    Cutmix & \multicolumn{3}{c}{1.0} \\
    Mixup & \multicolumn{3}{c}{0.8} \\
    Label smoothing & \multicolumn{3}{c}{0.1} \\
    EMA Decay Rate & \multicolumn{3}{c}{\xmark} \\
    Gradient Clipping  & \multicolumn{3}{c}{\xmark} \\
    Repeated Augmentation & \multicolumn{3}{c}{\xmark}  \\
   \bottomrule[1pt]
    \end{tabular}
    \label{tab:hyperparameter}.
\end{table}

\subsection{Datasets and Evaluation Metric}
We evaluate our method on three widely used benchmarks: ImageNet-1K, MS COCO 2017, and ADE20K datasets.
\begin{enumerate}
    \item Image classification: We use the ImageNet-1K dataset~\cite{deng2009imagenet1k}, which consists of 1.28 million training images and 50,000 validation images across 1,000 object categories.
    \item  Object detection and instance segmentation: Experiments are conducted on the MS COCO 2017 dataset~\cite{lin2014coco}, which contains around 118,000 training images and 5,000 validation images with bounding-box and instance-level annotations for 80 common categories.
    \item Semantic segmentation: We use the ADE20K dataset~\cite{zhou2019semantic} for the semantic segmentation task, which consists of 20,210 training images and 2,000 validation images with dense pixel-wise annotations over 150 semantic classes.
\end{enumerate} 
We evaluate model performance using standard metrics for each task and dataset. For image classification (ImageNet-1K), we report Top-1 validation accuracy as the primary evaluation metric. For object detection and instance segmentation (MS COCO dataset), we follow the standard COCO evaluation protocol and report mean Average Precision (mAP) metrics $\{AP, AP_{50}, AP_{75}\}$, where $AP_{50}$ and $AP_{75}$ correspond to IoU thresholds of $0.50$ and $0.75$, respectively. For semantic segmentation (ADE20K dataset), we evaluate performance using mean Intersection over Union (mIoU) metric.

\begin{table*}[t!]
\caption{Model performances on \textbf{ImageNet-1K Classification}. FLOPs are calculated with an input resolution of 224 $\times$ 224.
}
\begin{minipage}{0.5\linewidth}
\centering
\setlength{\tabcolsep}{0.5mm}
\resizebox{\linewidth}{!}{
\begin{tabular}{lc|ccc|c}
\toprule[1pt]
Method & Venue &  \begin{tabular}[c]{@{}c@{}} Img\\ Size\end{tabular}  & \#Param & FLOPs  & \begin{tabular}[c]{@{}c@{}}Top-1\\ Acc(\%)\end{tabular} \\
\midrule
RegNetY-4G~\cite{radosavovic2020regnety} & \multirow{3}{*}{\rotatebox{0}{\footnotesize cvpr'20 }} & \footnotesize 224$^2$ & 21M & 4.0G & 80.0 \\
RegNetY-8G~\cite{radosavovic2020regnety} &   & \footnotesize 224$^2$ & 39M & 8.0G  & 81.7 \\
RegNetY-16G~\cite{radosavovic2020regnety} &  & \footnotesize 224$^2$ & 84M & 16.0G  & 82.9 \\
\hline
ConvNeXt-T~\cite{liu2022convnet} &  \multirow{3}{*}{\rotatebox{0}{\footnotesize cvpr'22 }} & \footnotesize 224$^2$ & 29M & 4.5G & 82.1 \\ 
ConvNeXt-S~\cite{liu2022convnet}& & \footnotesize 224$^2$ & 50M & 8.7G & 83.1 \\ 
ConvNeXt-B~\cite{liu2022convnet} & & \footnotesize 224$^2$ & 89M & 15.4G & 83.8 \\ 
\hline
ViT-B/16~\cite{dosovitskiy2020vit}&  \multirow{2}{*}{\rotatebox{0}{\footnotesize iclr'21 }}  & \footnotesize 384$^2$ & 86M & 55.4G  & 77.9 \\
ViT-L/16~\cite{dosovitskiy2020vit} & & \footnotesize 384$^2$ & 307M & 190.7G  & 76.5 \\
\hline
DeiT-S~\cite{touvron2021deit} & \multirow{3}{*}{\rotatebox{0}{\footnotesize pmlr'21 }} & \footnotesize 224$^2$ & 22M & 4.6G  & 79.8 \\
DeiT-B~\cite{touvron2021deit} & & \footnotesize 224$^2$ & 86M & 17.5G  & 81.8 \\
DeiT-B~\cite{touvron2021deit} & & \footnotesize 384$^2$ & 86M & 55.4G  & 83.1 \\
\hline
Swin-T~\cite{liu2021swin} & \multirow{3}{*}{\rotatebox{0}{\footnotesize iccv'21 }} & \footnotesize 224$^2$ & 29M & 4.5G  & 81.3 \\
Swin-S~\cite{liu2021swin} & & \footnotesize 224$^2$ & 50M & 8.7G  & 83.0 \\
Swin-B~\cite{liu2021swin} & & \footnotesize 224$^2$ & 88M & 15.4G  & 83.5 \\
\hline
LIT-S~\cite{pan2022less} & \multirow{3}{*}{\rotatebox{0}{\footnotesize aaai'22 }} & \footnotesize 224$^2$ & 27M & 5.0G  & 81.5 \\
LIT-M~\cite{pan2022less} & & \footnotesize 224$^2$ & 48M & 8.6G  & 83.0 \\
LIT-B~\cite{pan2022less} & & \footnotesize 224$^2$ & 86M & 15.0G  & 83.4 \\
\hline
SLaK-T~\cite{liu2022more} & \multirow{3}{*}{\rotatebox{0}{\footnotesize iclr'23 }} & \footnotesize 224$^2$ & 30M & 5.0G  & 82.5 \\
SLaK-S~\cite{liu2022more} & & \footnotesize 224$^2$ & 55M & 9.8G  & 83.8 \\
SLaK-B~\cite{liu2022more} & & \footnotesize 224$^2$ & 95M & 17.1G  & 84.0 \\
\hline
InceptionNeXt-T~\cite{Yu2024inceptionnext} & \multirow{3}{*}{\rotatebox{0}{\footnotesize cvpr'24 }} & \footnotesize 224$^2$ & 28M & 4.2G  & 82.3 \\
InceptionNeXt-S~\cite{Yu2024inceptionnext} & & \footnotesize 224$^2$ & 49M & 8.4G  & 83.5 \\
InceptionNeXt-B~\cite{Yu2024inceptionnext} & & \footnotesize 224$^2$ & 87M & 14.9G  & 84.0 \\
\hline
S4ND-ConvNeXt-T~\cite{nguyen2022s4nd} &  \multirow{2}{*}{\rotatebox{0}{\scriptsize neurips22 }} & \footnotesize 224$^2$ & 30M & 5.2G & 82.2 \\
S4ND-ViT-B~\cite{nguyen2022s4nd} & & \footnotesize 224$^2$ & 88.8M & 17.1G & 80.4 \\
\hline
Mamba-2D-S~\cite{li2024mamba} &  \multirow{2}{*}{\rotatebox{0}{\footnotesize eccv'24 }} & \footnotesize 224$^2$ & 24M & - & 81.7 \\
Mamba-2D-B~\cite{li2024mamba} & & \footnotesize 224$^2$ & 92M & - & 83.0 \\
\hline
ViM-T~\cite{vim} &  \multirow{2}{*}{\rotatebox{0}{\footnotesize icml'24 }} & \footnotesize 224$^2$ & 7M & 1.5G & 76.1 \\
ViM-S~\cite{vim} & & \footnotesize 224$^2$ & 26M & 5.1G & 80.5 \\
\bottomrule[1pt]
\end{tabular}
}
\end{minipage}
\begin{minipage}{0.5\textwidth}
\centering
\setlength{\tabcolsep}{0.5mm}
\resizebox{\linewidth}{!}{
\begin{tabular}{lc|ccc|c}
\toprule[1pt]
Method & Venue &  \begin{tabular}[c]{@{}c@{}} Img\\ Size\end{tabular}  & \#Param & FLOPs  & \begin{tabular}[c]{@{}c@{}}Top-1\\ Acc(\%)\end{tabular} \\
\midrule
VMamba-T~\cite{vmamba}& \multirow{3}{*}{\rotatebox{0}{\scriptsize neurips24 }} & \footnotesize 224$^2$ & 30M & 4.9G & 82.6 \\ 
VMamba-S~\cite{vmamba}& & \footnotesize 224$^2$ & 50M & 8.7G & 83.6 \\ 
VMamba-B~\cite{vmamba}& & \footnotesize 224$^2$ & 89M & 15.4G & 83.9 \\ 
\hline
PlainMamba-L1~\cite{yang2024plainmamba}&\multirow{3}{*}{\rotatebox{0}{\footnotesize bmvc'24 }} & \footnotesize 224$^2$ & 7M & 3.0G & 77.9 \\ 
PlainMamba-L2~\cite{yang2024plainmamba}& & \footnotesize 224$^2$ & 25M & 8.1G & 81.6 \\ 
PlainMamba-L3~\cite{yang2024plainmamba} & & \footnotesize 224$^2$ & 50M & 14.4G & 82.3 \\ 
\hline
LocalVMamba-T~\cite{huang2024localmamba} &  \multirow{2}{*}{\rotatebox{0}{\footnotesize eccv'24 }} & \footnotesize 224$^2$ & 26M & 5.7G & 82.7 \\
LocalVMamba-S~\cite{huang2024localmamba} & &  \footnotesize 224$^2$ & 50M & 11.4G & 83.7 \\
\hline
MSVMamba-N~\cite{shi2024multi}& \multirow{3}{*}{\rotatebox{0}{\scriptsize neurips24 }} & \footnotesize 224$^2$ & 7M & 0.9G & 77.3 \\ 
MSVMamba-M~\cite{shi2024multi}& & \footnotesize 224$^2$ & 12M & 1.5G & 79.8 \\ 
MSVMamba-T~\cite{shi2024multi}& & \footnotesize 224$^2$ & 33M & 4.6G & 82.8 \\ 
\hline
VRWKV-T~\cite{duan2025visionrwkv}& \multirow{3}{*}{\rotatebox{0}{\footnotesize iclr'25 }} & \footnotesize 224$^2$ & 6.2M & 1.2G & 75.1 \\ 
VRWKV-S~\cite{duan2025visionrwkv}& & \footnotesize 224$^2$ & 23.8M & 4.6G & 80.1 \\ 
VRWKV-B~\cite{duan2025visionrwkv}& & \footnotesize 224$^2$ & 93.7M & 18.2G & 82.0 \\ 
\hline
FractalMamba-T~\cite{Xiao2025fractablmamba}& \multirow{1}{*}{\rotatebox{0}{\footnotesize aaai'25 }} & \footnotesize 224$^2$ & 31M & 4.8G & 83.0 \\ 
\hline
EffVMamba-T~\cite{Pei2025EfficientVMamba}& \multirow{3}{*}{\rotatebox{0}{\footnotesize aaai'25 }} & \footnotesize 224$^2$ & 6M & 0.8G & 76.5 \\ 
EffVMamba-S~\cite{Pei2025EfficientVMamba}& & \footnotesize 224$^2$ & 11M & 1.3G & 78.7 \\ 
EffVMamba-B~\cite{Pei2025EfficientVMamba}& & \footnotesize 224$^2$ & 33M & 4.0G & 81.8 \\ 
\hline
MambaVision-T1~\cite{hatamizadeh2025mambavision}&\multirow{4}{*}{\rotatebox{0}{\footnotesize cvpr'25 }} & \footnotesize 224$^2$ & 31.8M & 4.4G & 82.3 \\ 
MambaVision-T2~\cite{hatamizadeh2025mambavision}& & \footnotesize 224$^2$ & 35.1M & 5.1G & 82.7 \\ 
MambaVision-S~\cite{hatamizadeh2025mambavision}& & \footnotesize 224$^2$ & 50.1M & 7.5G & 83.3 \\ 
MambaVision-B~\cite{hatamizadeh2025mambavision}& & \footnotesize 224$^2$ & 97.7M & 15.0G & 84.2 \\
\hline
Mamba®-T~\cite{wang2025mambar}& \multirow{3}{*}{\rotatebox{0}{\footnotesize cvpr'25 }} & \footnotesize 224$^2$ & 9M & 5.1G & 77.4 \\ 
Mamba®-S~\cite{wang2025mambar}& & \footnotesize 224$^2$ & 28M & 9.9G & 81.4 \\ 
Mamba®-B~\cite{wang2025mambar}& & \footnotesize 224$^2$ & 99M & 20.3G & 83.0 \\ 
\hline
\rowcolor{customrowcolor}
MFilMamba-T& & \footnotesize 224$^2$ & 33.5M & 5.6G & 83.2 \\ 
\rowcolor{customrowcolor}
MFilMamba-S& \footnotesize  & \footnotesize 224$^2$ & 50.6M & 9.1G & 83.9  \\ 
\rowcolor{customrowcolor}
MFilMamba-B & & \footnotesize 224$^2$ & 93.1M  & 16.8G & 84.2 \\
\bottomrule[1pt]
\end{tabular}
}
\end{minipage}
\label{table:classification}
\end{table*}

\subsection{Image Classification}
\label{subsec:classification}
\noindent {\em Settings.} We conduct image classification experiments on the ImageNet-1K dataset~\cite{deng2009imagenet1k}. Each model variant is trained for 300 epochs with a batch size of 1024 on 8 NVIDIA A100 GPUs. Both training and inference are conducted at a resolution of \(224^{2}\). Following~\cite{touvron2021deit}, we apply data augmentation and regularization including RandAugment~\cite{Cubuk2020randaugment}, MixUp~\cite{zhang2017mixup}, CutMix~\cite{yun2019cutmix}, Random Erasing~\cite{zhong2020randerasing}, and Stochastic Depth~\cite{Huang2016stochasticdepth}. AdamW~\cite{kingma2014adam} optimizer is used with a weight decay of 0.05 and a cosine-annealed learning rate starting at 0.001. We set the stochastic drop path rate to 0.2 for the Tiny variant, 0.3 for the Small variant, and 0.5 for the Base variant. The complete hyperparameter configuration of our models used for the ImageNet-1K classification tasks is shown in Table~\ref{tab:hyperparameter}.

\smallskip
\noindent {\em Results analysis.} Table~\ref{table:classification} compares MFil-Mamba models with existing state-of-the-art models based on convolutional, transformer, and SSM architectures. The tiny variant, MFil-Mamba-T, achieves 83.2\% Top-1 validation accuracy on ImageNet-1k, surpassing Swin-T by 1.9\%, RegNetY-8G by 1.5\%, ConvNeXt-T by 1.1\%, SLaK-T by 0.7\% and InceptionNeXt-T by 0.9\%. Compared to recent SSM-based models with similar parameters and FLOPs, MFil-Mamba-T outperforms VMamba-T by 0.6\%, PlainMamba-L2 by 1.6\%, EfficientVMamba-B by 1.4\%, Mamba®-S by 1.8\%, and Mamba-Vision-T1/T2 by 0.9\%/0.5\%. The small variant, MFil-Mamba-S, achieves 83.9\% Top-1 accuracy, surpassing existing counterparts of similar size and complexity. Similarly, the base variant, MFil-Mamba-B, achieves 84.2\% top-1 accuracy, outperforming existing baselines including Swin-B by 0.7\%, ConvNeXt-B by 0.4\%, Mamba-2D-B by 1.2\%,  VMamba-B by 0.3\%, and Mamba®-B by 1.2\%. These results demonstrate the strong performance and scalability of MFil-Mamba across different model sizes.

\subsection{Object Detection and Instance Segmentation}
\label{subsec:det}

\begin{table*}[tbp]
    \centering
       \caption{\textbf{Object detection and instance segmentation} performance on \textbf{MS COCO} dataset~\cite{lin2014coco} using the Mask RCNN framework~\cite{he2017maskrcnn} with 1$\times$ schedule. FLOPs are calculated with an input resolution of 1280 $\times$ 800. AP$^{b}$ and AP$^{m}$ represents mean average precision (AP) of bounding box and mask, respectively.}
    \setlength{\tabcolsep}{3mm}
    \resizebox{0.97\linewidth}{!}{
    \begin{tabular}{l|cc|ccc||ccc}
        \toprule[1pt]
        \multirow{2}{*}{Backbone} &  \multirow{2}{*}{\#Param} & \multirow{2}{*}{FLOPs} & \multicolumn{3}{c||}{Object Detection} & \multicolumn{3}{c}{Instance Segmentation}  \\
        \cline{4-9}
        
        & & & AP$^{\rm box}$ & AP$_{50}^{\rm box}$ &AP$_{75}^{\rm box}$ & AP$^{\rm mask}$ & AP$_{50}^{\rm mask}$ &AP$_{75}^{\rm mask}$ \\
        \hline

        Swin-T~\cite{liu2021swin} & 48M & 267G & 42.7 & 65.2 & 46.8 & 39.3 & 62.2 & 42.2\\
        ConvNeXt-T~\cite{liu2022convnet} &  48M & 262G & 44.2 & 66.6 & 48.3 & 40.1 & 63.3 & 42.8 \\
         PVT-S~\cite{wang2021pyramid} & 44M &  245G & 40.4 &  62.9 & 43.8 &  37.8 & 60.1 & 40.3 \\
        DAT-T~\cite{xia2022vision} & 48M & 272G  & 44.4 & 67.6 & 48.5  & 40.4  & 64.2  & 43.1  \\
        ViT-Adapter-S~\cite{chen2022vitadapter} & 48M  &  403G & 44.7  & 65.8 & 48.3 & 39.9  & 62.5  & 42.8 \\
        VMamba-T~\cite{vmamba} &  50M & 271G & 47.3  & 69.3 & 52.0 & 42.7 & 66.4 & 45.9\\ 
        PlainMamba-L1~\cite{yang2024plainmamba} &  31M & 388G & 44.1 &  64.8 & 47.9 & 39.1 & 61.6 & 41.9\\
        MambaOut-Tiny~\cite{yu2025mambaout} &  43M & 262G & 45.1 & 67.3 & 49.6 &  41.0 & 64.1   & 44.1 \\
        EffVMamba-B~\cite{Pei2025EfficientVMamba} &  53M & 252G & 43.7 & 66.2 & 47.9  & 40.2 & 63.3 & 42.9 \\
        FractalMamba-T~\cite{Xiao2025fractablmamba} &  41M & 266G & 46.8 & 68.7 & 50.8  & 42.4 & 65.9 & 45.8 \\
        \rowcolor{customrowcolor}
        MFilMamba-T&  53M & 261G &  47.3 &  69.2 &  51.9 & 42.7 &  66.2 & 46.0 \\
        
        \hline
        Swin-S~\cite{liu2021swin} & 69M & 354G & 44.8 & 66.6 & 48.9 & 40.9 & 63.2 & 44.2\\
        ConvNeXt-S~\cite{liu2022convnet} &  70M & 348G & 45.4 & 67.9 & 50.0 & 41.8 & 65.2 & 45.1\\
         PVT-M~\cite{wang2021pyramid} &  64M & 302G & 42.0 & 64.4  & 45.6 & 39.0 & 61.6 & 42.1 \\
        DAT-S~\cite{xia2022vision} & 69M & 378G & 47.1 & 69.9 & 51.5  & 42.5 & 66.7  & 45.4  \\
        VMamba-S~\cite{vmamba} &  70M & 349G & 48.7 & 70.0 & 53.4 & 43.7 & 67.3  & 47.0 \\
        PlainMamba-L2~\cite{yang2024plainmamba} &  53M & 542G & 46.0 & 66.9 & 50.1 &  40.6 & 63.8  & 43.6 \\
        MambaOut-Small~\cite{yu2025mambaout} &  65M &  354G & 47.4 & 69.1 & 52.4 & 42.7 & 66.1 & 46.2 \\
        \rowcolor{customrowcolor}
        MFilMamba-S &  70M & 320G & 47.9 &  69.4 & 52.3  & 43.1 & 66.8  & 46.4 \\

        \hline
        Swin-B~\cite{liu2021swin} & 107M & 496G & 46.9 & - & - & 42.3 & - & - \\
        ConvNeXt-B~\cite{liu2022convnet} &  108M & 486G & 47.0 & 69.4 & 51.7 & 42.7 & 66.3 & 46.0\\  
         PVT-L~\cite{wang2021pyramid} & 81M  & 364G & 42.9 & 65.0 & 46.6 & 39.5 & 61.9 & 42.5 \\
        ViT-Adapter-B~\cite{chen2022vitadapter} & 120M  & 557G  & 47.0  & 68.2 & 51.4 & 41.8  & 65.1  & 44.9 \\
        VMamba-B~\cite{vmamba}&  108M & 485G & 49.2 & 71.4 & 54.0 & 44.1 & 68.3 & 47.7  \\
        PlainMamba-L3~\cite{yang2024plainmamba}&  79M & 696G &  46.8 & 68.0  & 51.1  & 41.2 &  64.7 & 43.9 \\
        MambaOut-Base~\cite{yu2025mambaout} & 100M  &  495G & 47.4 & 69.3 & 52.2 &  43.0 &  66.4  & 46.3 \\
        \rowcolor{customrowcolor}
        MFilMamba-B& 112M & 467G & 49.0 & 70.6 & 53.7  & 43.8 & 67.7  & 47.6\\
        \bottomrule[1pt]
    \end{tabular}
    }
    \label{table:det-seg}
\end{table*}

\noindent {\em Settings.} We evaluate our model on MS COCO dataset~\cite{lin2014coco} for object detection and instance segmentation tasks using Mask R-CNN~\cite{he2017maskrcnn} and MMDetection library~\cite{chen2019mmdet}. We initialize MFil-Mamba backbone with ImageNet-1K~\cite{deng2009imagenet1k} pretraining and fine-tune it for 12 epochs (1\(\times\) training schedule) with a batch size of 16. We use the AdamW~\cite{kingma2014adam} optimizer with an initial learning rate set at 0.0001.

\begin{figure}[tbp]
    \centering  
    \includegraphics[width=\linewidth]{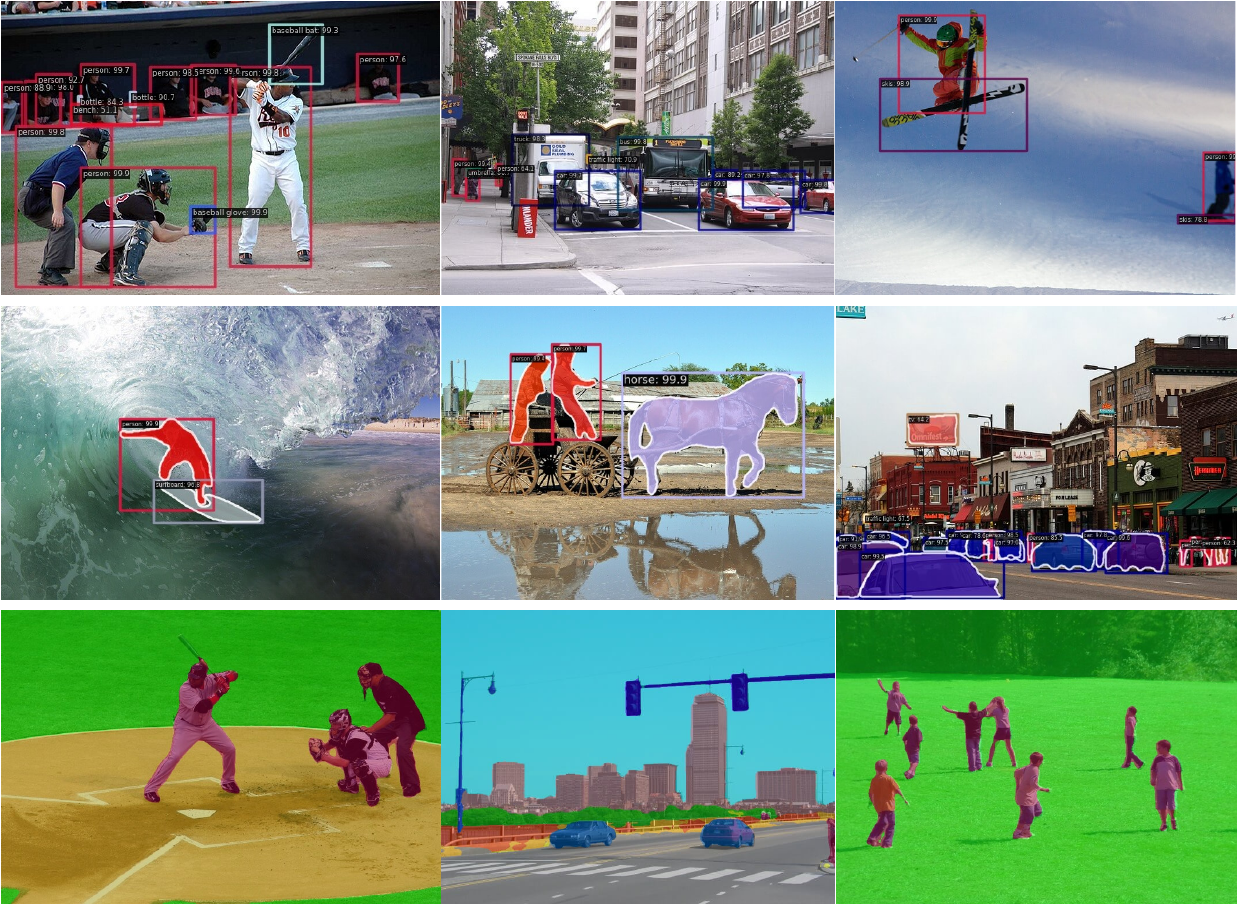}
    \caption{Qualitative visualizations of MFil-Mamba-T for object detection (1st row) and instance segmentation tasks (2nd row) on the MS COCO dataset~\cite{lin2014coco}, and for semantic segmentation task (3rd row) on the ADE20K dataset~\cite{zhou2019semantic}. Detailed visualization is presented in Sec~\ref{sec:qualitative_results}.}
    \label{fig:qualitative_viz}
\end{figure}

\smallskip
\noindent {\em Results analysis.}  Table~\ref{table:det-seg} compares MFil-Mamba with other models under a 1$\times$ training schedule. Notably, MFil-Mamba-T achieves 47.3 box AP and 42.7 mask AP, surpassing Swin-T, PlainMamba-L1, MambaOut-Tiny, and EfficientVMamba-B by +4.6, +3.2, +2.2, and +3.6 box AP, and by +3.4, +3.6, +1.7, and +2.5 mask AP, respectively, and is comparable to VMamba-T. Similarly, MFil-Mamba-S achieves 47.9 box AP and 43.1 mask AP, consistently outperforming its counterparts. MFil-Mamba-B attains 49.0 box AP and 43.8 mask AP, outperforming backbones with similar parameters and FLOPs. For instance, it surpasses ViT-Adapter-B, PlainMamba-L3, and MambaOut-Base by +2.0, +2.2, and +1.6 box AP, and by +2.0, +2.6, and +0.8 mask AP, respectively.  Compared to VMamba Small and Base, the results are slightly lower, but overall MFil-Mamba demonstrates strong efficiency and generalizability across the object detection and instance segmentation tasks. To further illustrate the performance of our model, qualitative visualization results on the MS COCO~\cite{lin2014coco} validation set are shown in Fig.~\ref{fig:qualitative_viz}.

\begin{table}[!t]
\centering
\small
   \caption{\textbf{Semantic segmentation performance} of different backbones on \textbf{ADE20K}~\cite{zhou2019semantic} using UPerNet~\cite{xiao2018upernet}. FLOPs are computed for an input resolution of 512 $\times$ 2048. SS and MS denote single-scale and multi-scale evaluation, respectively.}
\resizebox{\linewidth}{!}{
\begin{tabular}{l|cc|cc}
\toprule[1pt]
Backbone & \#Param & FLOPs &  \makecell{mIoU\\(SS)}  & \makecell{mIoU\\(MS)}\\
\midrule
ResNet-50~\cite{he2016resnet}& 67M & 953G & 42.1 & 42.8 \\
Swin-T~\cite{liu2021swin} & 60M & 945G & 44.4 & 45.8\\
Focal-T~\cite{yang2021focal} & 62M & 998G & 45.8 & 47.0 \\
DAT-T~\cite{xia2022vision} &  60M & 957G & 45.5  & 46.4 \\
ConvNeXT-T~\cite{liu2022convnet} & 60M & 939G & 46.0 & 46.7 \\
NAT-T~\cite{hassani2023neighborhood} & 58M & 934G &  47.1 & 48.4 \\
VMamba-T~\cite{vmamba} & 62M & 949G & 47.9 & 48.8\\
PlainMamba-L2~\cite{yang2024plainmamba} & 55M & 285G & 46.8 & - \\
MambaVision-T~\cite{hatamizadeh2025mambavision} & 55M & 945G & 46.0 & - \\
EffVMamba-B~\cite{Pei2025EfficientVMamba} & 65M & 930G & 46.5 & 47.3 \\
\rowcolor{customrowcolor}
MFilMamba-T & 64M & 961G & 48.5 & 48.9\\

\hline
ResNet-101~\cite{he2016resnet}& 86M & 1030G & 43.8 & 44.9 \\
Swin-S~\cite{liu2021swin}& 81M & 1039G & 47.6 & 49.5 \\
Focal-S~\cite{yang2021focal} & 85M & 1130G & 48.0 & 50.0 \\
DAT-S~\cite{xia2022vision} & 81M  &  1079G & 48.3 & 49.8  \\
ConvNeXt-S~\cite{liu2022convnet} & 82M & 1027G & 48.7 & 49.6 \\
NAT-S~\cite{hassani2023neighborhood} & 82M & 1010G &  48.0 & 49.5 \\
VMamba-S~\cite{vmamba}& 82M & 1028G & 50.6 & 51.2 \\
MambaVision-S~\cite{hatamizadeh2025mambavision} & 84M & 1135G & 48.2 & - \\
\rowcolor{customrowcolor}
MFilMamba-S & 81M & 1028G & 49.3 & 49.9\\

\hline
Swin-B~\cite{liu2021swin} & 121M & 1188G & 48.1 &  49.7 \\
Focal-B~\cite{yang2021focal} & 126M & 1354G & 49.0 &  50.5\\
DAT-B~\cite{xia2022vision} & 121M & 1212G & 49.4 & 50.6  \\
ConvNeXt-B~\cite{liu2022convnet} & 122M & 1170G & 49.1 & 49.9 \\
NAT-B~\cite{hassani2023neighborhood} & 123M & 1137G &  48.5 & 49.7 \\
VMamba-B~\cite{vmamba} & 122M & 1170G & 51.0 & 51.6 \\
MambaVision-B~\cite{hatamizadeh2025mambavision} & 126M & 1342G & 49.1 & - \\
\rowcolor{customrowcolor}
MFilMamba-B & 125M & 1198G & 50.6 & 51.3 \\
\bottomrule[1pt]
\end{tabular}
}
 
\label{table:seg}
\end{table}
\subsection{Semantic Segmentation}
\label{subsec:semantic-seg}
\noindent {\em Settings.} We conduct semantic segmentation experiments on the ADE20K dataset~\cite{zhou2019semantic} using UPerNet~\cite{xiao2018upernet} with the MMSegmentation library~\cite{2020mmseg}. The backbone is initialized with weights pretrained on ImageNet-1K~\cite{deng2009imagenet1k} and fine-tuned for 160k iterations with a batch size of 16. We employ the AdamW~\cite{kingma2014adam} optimizer with an initial learning rate of $6\times10^{-5}$. We conduct experiments using the default input resolution of 512 × 512. Both single-scale (SS) and multi-scale (MS) inference results are reported in Table~\ref{table:seg}.

\smallskip
\noindent {\em Results analysis.} Table~\ref{table:seg} compares the MFil-Mamba backbones with existing models on the semantic segmentation task. MFil-Mamba-T achieves 48.5 mIoU (SS) and 48.9 mIoU (MS), surpassing Swin-T, ConvNeXT-T, EfficientVMamba-T, and VMamba-T by +4.1, +2.5, +2.0, +0.6 (SS), and by +3.1,  +2.2, +1.6, +0.1 (MS), respectively. It also surpasses MambaVision-T and PlainMamba-L2 by +2.5 and +1.7 mIoU (SS). Our small and base variants further outperform existing models of similar size and complexity. Specifically, MFil-Mamba-S gains +1.7, +1.3, and +1.1 mIoU (SS) over Swin-S, Focal-S, and MambaVision-S, while MFil-Mamba-B improves by +2.5, +1.6, and +1.5 mIoU (SS) over Swin-B, Focal-B, and MambaVision-B, respectively. Although slightly below the VMamba small and base variants, MFil-Mamba consistently outperforms other latest Transformer-based and SSM-based counterparts. 

Qualitative results on the ADE20K dataset~\cite{zhou2019semantic} are shown in Fig.~\ref{fig:qualitative_viz} (third row). These results demonstrate the effectiveness and significant potential of the MFil-Mamba architecture for semantic segmentation and other related dense prediction tasks.

\begin{figure}[t!]
  \centering
  \includegraphics[width=0.97\linewidth]{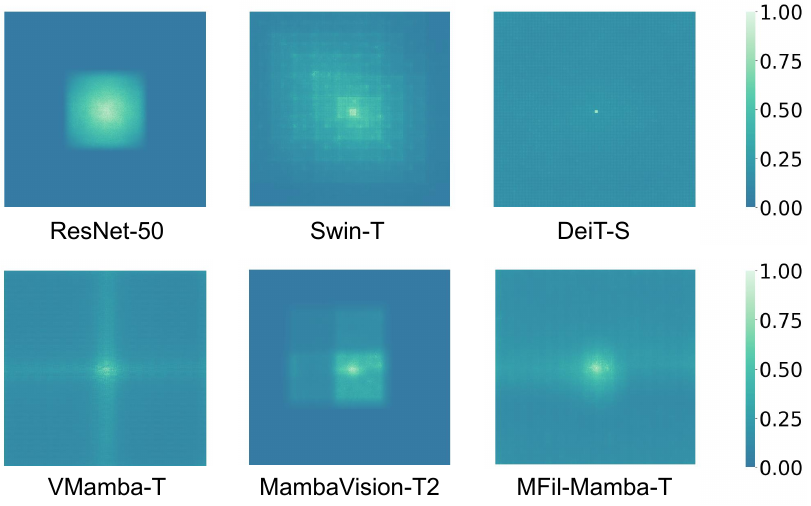}
  \caption{Comparison of MFil-Mamba’s effective receptive field with existing state-of-the-art baselines.}
  \label{fig:erf}
\end{figure}

\subsection{Effective Receptive Field (ERF) Visualization}
\label{subsec:erf}
Fig.~\ref{fig:erf} presents ERF~\cite{luo2016erf} visualizations for different architectures including Convolutional, Transformers, State Space Model and Hybrid. It highlights which regions of the input most influence the model output. The color gradient from dark to light green in the diagram indicates increasing contribution to the receptive field. Dark green represents a more localized focus, and as the color moves toward lighter green, it reflects a greater contribution to global dependencies. ResNet-50 shows a highly localized ERF, consistent with CNN feature extraction. DeiT-S exhibits a centralized global ERF, while Swin-T maintains locality due to window-based self-attention. VMamba-T achieves a global receptive field but with cross-like artifacts resulting from directional traversal. Mamba-Vision-T2 shows a mostly local receptive field. In contrast, our MFil-Mamba-T demonstrates a strong central focus with a smooth global receptive field, free from unnatural patterns. Furthermore, this global and dense receptive field, which is an intrinsic property of state-space models, highlights that the observed performance improvements do not arise from the inductive bias of filter. Rather, they arise from the SSM operating on a more coherent and structurally enriched representation of the input features, which is further supported by theoretical justification (Section~\ref{subsec:theory}).

\begin{figure*}[tbp]
    \centering
    \includegraphics[width=\linewidth]{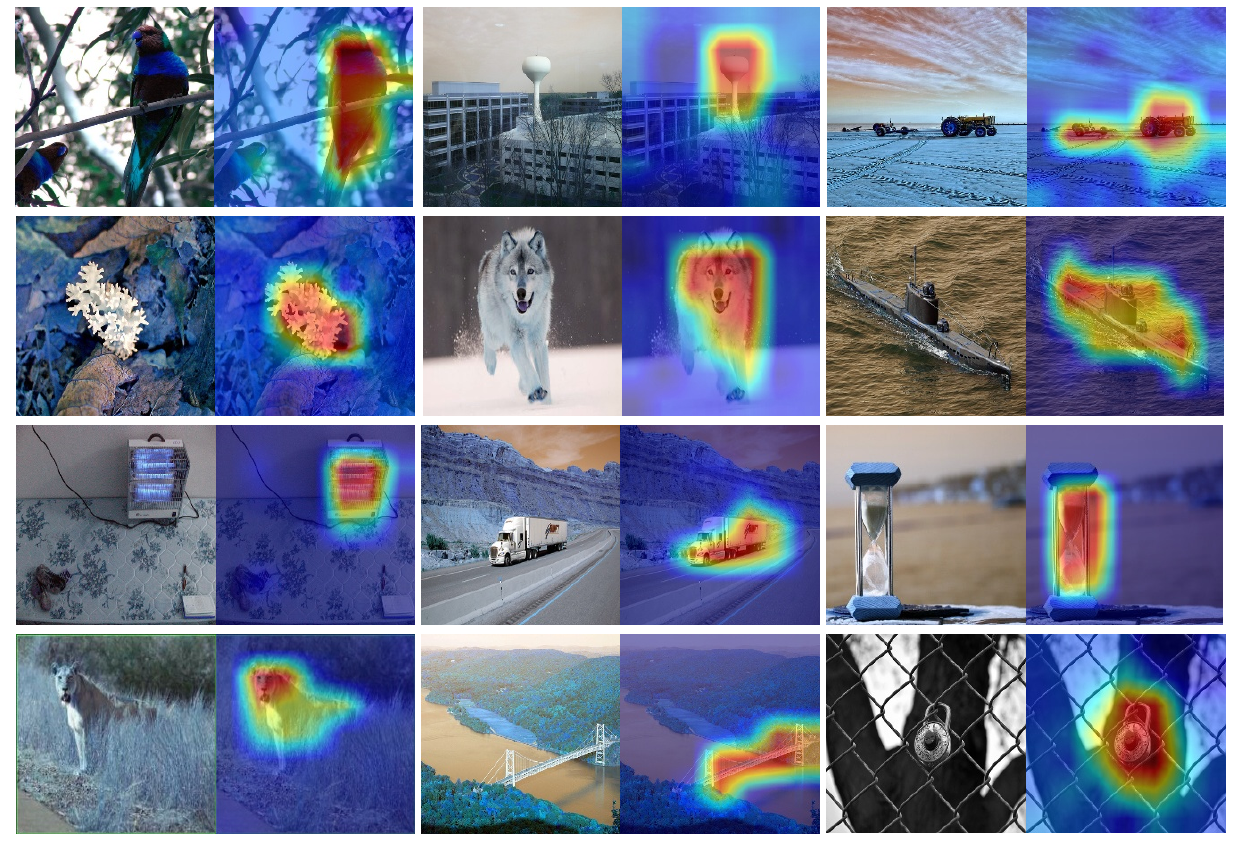}
    \caption{\textbf{Grad-CAM visualization} on the \textbf{ImageNet-1k} validation set, illustrating model's ability to focus on the main subject of the image.}
    \label{fig:viz_gradcam}
\end{figure*}

\subsection{Model Interpretability}
\label{sec:interpretability}
Several visualization tools have been developed for model interpretation and to better understand its internal reasoning process~\cite{Selvaraju2017gradcam, wall2026winsor, Chattopadhay2018gradcamp}. For our model, we compute Grad-CAM visualizations on samples from the ImageNet-1k validation set, as shown in Fig.~\ref{fig:viz_gradcam}. We use the Tiny variant, MFil-Mamba-T, to visualize the gradient-based activation map in each image. From the visualizations, it is evident that our model is able to accurately focus on the main subject of each image and hence highlight the regions that are most relevant to the predicted class. Specifically, the visualizations remain sharp and coherent across various object categories, including birds, animals, vehicles, and more. This  suggests that the model has successfully learned to attend to semantically meaningful components of the image, rather than being distracted by irrelevant background information.

\begin{figure*}[tbp]
    \centering
    \includegraphics[width=\linewidth]{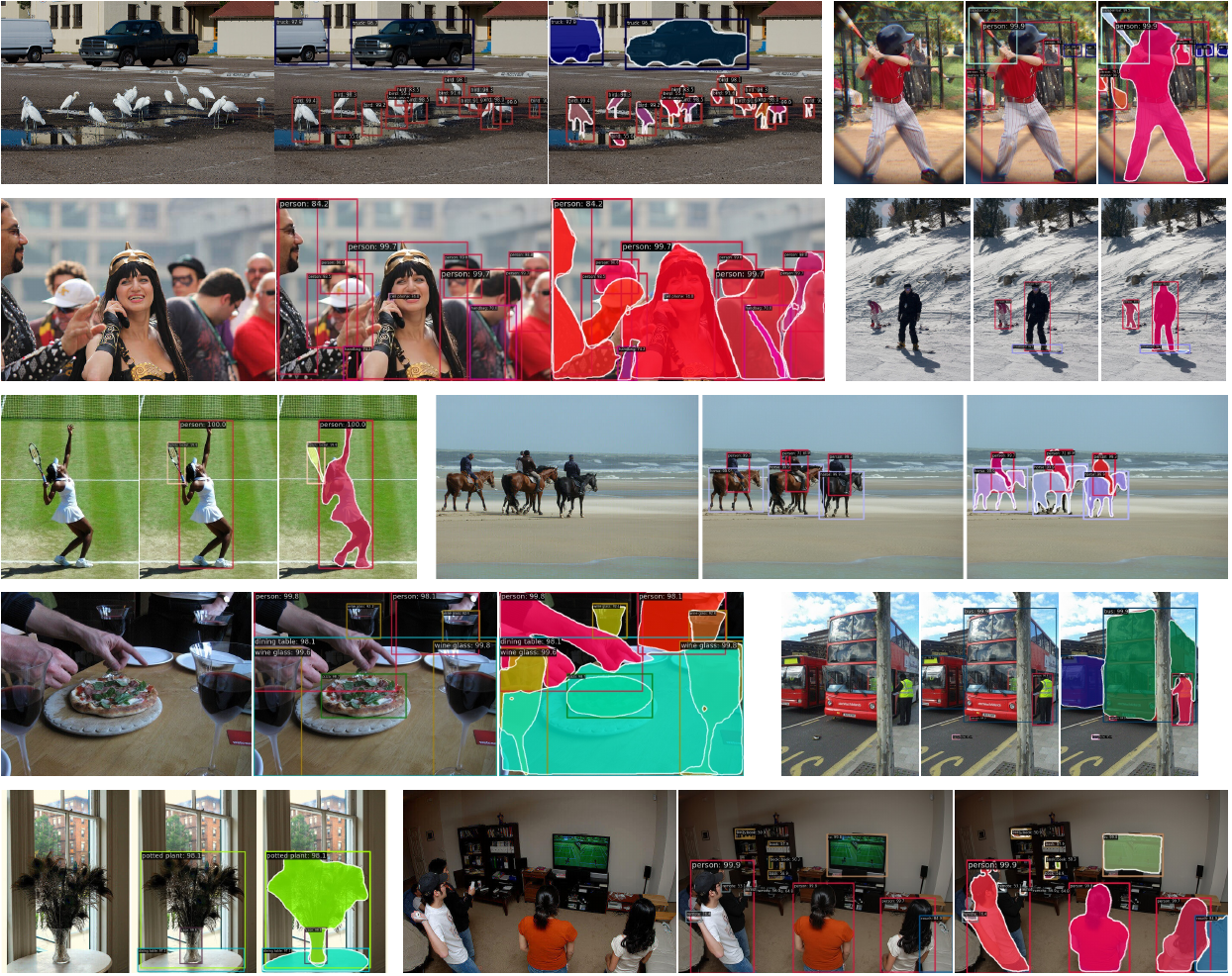}
    \caption{\textbf{Object detection and instance segmentation visualizations} on the \textbf{MS COCO} validation set. In each group of three successive images, the left image shows the original input, the middle shows object detection results with bounding boxes, and the right shows instance segmentation results with masks and bounding boxes. }
    \label{fig:viz_det_inst_seg}
\end{figure*}

\begin{figure*}[tbp]
    \centering
    \includegraphics[width=0.982\linewidth]{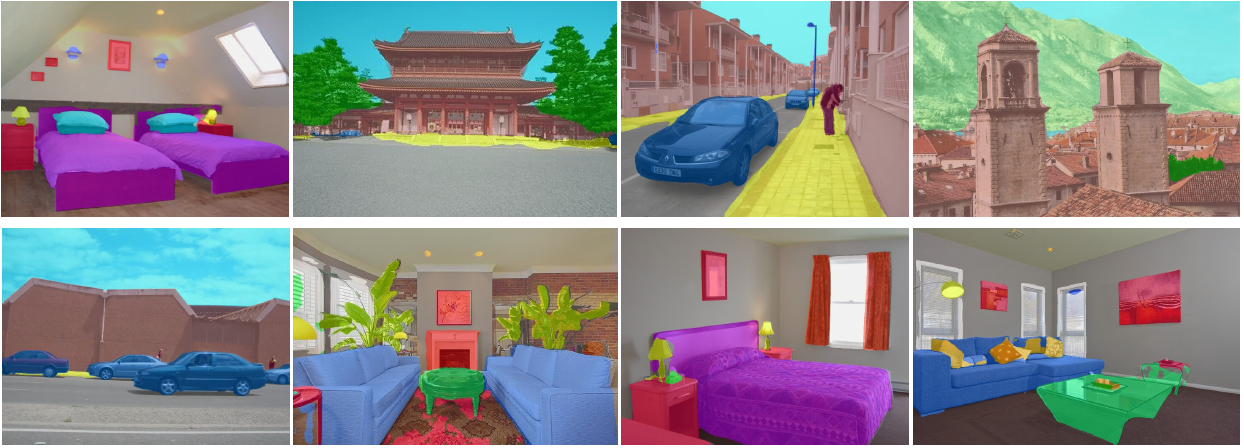}
    \caption{\textbf{Semantic segmentation visualizations} on the \textbf{ADE20k} dataset.}
    \label{fig:viz_sema_seg}
\end{figure*}

\subsection{Qualitative Visualizations}
\label{sec:qualitative_results}
Fig.~\ref{fig:viz_det_inst_seg} shows qualitative visualizations for object detection and instance segmentation on the MS COCO validation dataset using our tiny variant, MFilMamba-T. Each group in Fig.~\ref{fig:viz_det_inst_seg} contains three images: the first represents original input image, the second shows object detection results with bounding boxes, and the third shows the predicted instance masks along with labels for the instance segmentation task. As observed in the Fig.~\ref{fig:viz_det_inst_seg}, our model generates accurate bounding boxes around objects with high precision and correctly identifies instance masks. 

Similarly, Fig.~\ref{fig:viz_sema_seg} presents qualitative results for semantic segmentation on the ADE20K dataset. The semantic segmentation outputs further highlight the capability of our model to produce precise and coherent segmentation maps ranging from simple to complex scene. These qualitative visualizations across diverse and challenging datasets demonstrate the robustness and effectiveness of our model as a strong backbone for various downstream tasks, including object detection, instance segmentation, and semantic segmentation.

\subsection{Ablation Study}
We perform a series of ablation studies on MFil-Mamba-T to evaluate the efficiency and contribution of its individual components on the ImageNet-1K classification task. Specifically, we examine the effect of the filter-based scanning module, the adaptive weighting module, and different model configurations. We run each experiment for 300 epochs with the same hyperparameter configuration for the ablation. These studies aim to quantify how each component contributes to overall performance of the proposed model.

\smallskip
\noindent {\em a) Effect of Filter-Based Scan.} We evaluate the filter-based scanning module via ablation study (Table~\ref{tab:ablt_filter}) on ImageNet-1K. A single-directional scan using flattened image patches achieves 82.79\% Top-1 accuracy with 30.93M parameters and 4.94 GFLOPs. Replacing the flattening traversal with a four-directional cross-scan improves the accuracy to 82.95\% while maintaining comparable computational cost (5.14 GFLOPs). We then enhance the baseline by introducing a learnable filter that operates jointly with the original feature map and achieves 83.06\% Top-1 accuracy. Finally, we extend this design to three dynamic filters that capture vertical, horizontal, and complementary spatial cues, further increasing performance to 83.22\% Top-1 accuracy with 33.46M parameters and 5.56 GFLOPs. These results highlight the effectiveness and strong generalization ability of the multi-filter scanning module over standard directional traversal in visual state-space models.

\begin{table}[t]
    \centering
    \small
      \caption{Ablation of scanning modules in the Tiny variant.}
    \resizebox{\linewidth}{!}{
    \begin{tabular}{l|cc|c}
    \toprule
     Strategy  & \#Params & FLOPs & Top-1 Acc.  \\
       \midrule
       Single Scan (Flatten) & 30.93M & 4.94G  & 82.79 \\
    Cross Scan (4-Direction) & 30.93M  & 5.14G & 82.95 \\
     Original + Single Filter & 33.36M  & 5.41G & 83.06 \\
       Multi Filter Scan & 33.46M & 5.56G & 83.22\\
       \bottomrule   
    \end{tabular}
    }
    \label{tab:ablt_filter}
\end{table}
\smallskip

\noindent {\em b) Effect of Adaptive Weighting.} Table~\ref{tab:ablt_adpt_wt} presents the ablation results for the adaptive weighting module (Section~\ref{subsec:adaptive_merging}). We remove the lightweight adaptive weighting scheme from the baseline model and evaluate it by directly merging the four scan outputs. Without adaptive weighting, our tiny variant achieves 83.10\% Top-1 accuracy, showing a reduction of 0.12\% compared to the baseline model. Incorporating adaptive weighting introduces only a very small negligible increase in parameters while consistently improving performance. These results highlight the effectiveness of the proposed lightweight trainable adaptive weighting strategy for combining multiple scan outputs.

\smallskip
\noindent {\em c) Effect of Model Configuration.} We further investigate the impact of architectural design choices in MFil-Mamba by varying the state dimension ($\mathbf{d\_state}$) and the feature expansion ratio ($\mathbf{ssm\_ratio}$). These two hyperparameters determine the representational capacity of the model. Table~\ref{tab:ablat_config} summarizes the results under different configurations. When $\mathbf{ssm\_ratio}=2$, increasing $\mathbf{d\_state}$ from 1 to 2 provides a modest gain in Top-1 accuracy. However, the configuration with $\mathbf{d\_state}=1$ and $\mathbf{ssm\_ratio}=1$ achieves the highest Top-1 accuracy of 83.22\%, with only 33.46M parameters and 5.56 GFLOPs. These results indicate that simply increasing $\mathbf{d\_state}$ or $\mathbf{ssm\_ratio}$ does not necessarily lead to better performance. Instead, a balanced coordination of these hyperparameters is needed for stable optimization and effective feature interaction.

\begin{table}[tb]
    \centering
    \small
        \caption{Ablation study of adaptive weighting implementation.}
    \resizebox{\linewidth}{!}{
    \begin{tabular}{l|cc| c}
    \toprule
         Merging & \#Param & FLOPs & Top-1 Acc.  \\
         \midrule
         Without Adpt. Weighting & 33.46M & 5.56G & 83.10  \\
         With Adpt. Weighting & 33.46M & 5.56G & 83.22 \\
         \bottomrule
    \end{tabular}
    }
    \label{tab:ablt_adpt_wt}
\end{table}

\begin{table}[tb]
    \centering
    \small
      \caption{Ablation study on (different) model configurations.}
    \resizebox{\linewidth}{!}{
    \begin{tabular}{cc|cc|cc}
    \toprule
    $\mathbf{d\_state}$  & $\mathbf{ssm\_ratio}$ & \#Params & FLOPs & Top-1 Acc.  \\
   \midrule
   1 & 1.0 &  33.46M & 5.56G & 83.22\\
   1 & 2.0 & 48.27M & 8.21G & 83.08 \\
   2 & 2.0 & 48.31M & 8.34G & 83.10 \\
         \bottomrule
    \end{tabular}
    } 
    \label{tab:ablat_config}
\end{table}

%% file: sec/5_conclusion.tex
\section{Conclusion}
\label{sec:conclusion}
In this paper, we have introduced MFil-Mamba, a novel multi-filter scanning architecture designed to address the limitations of pre-defined directional traversal in existing Mamba-based vision models. By incorporating a multi-filter scanning backbone, adaptive weighting, and optimized architectural configurations, our approach substantially enhanced representational capacity while effectively mitigating the sequential processing constraints of standard Mamba architectures. Extensive experiments on ImageNet-1K and multiple downstream tasks, such as object detection, instance segmentation, and semantic segmentation, demonstrated that MFil-Mamba consistently outperforms both Transformer-based and SSM-based counterparts. These results proved the effectiveness, scalability, and generalizability of our architecture, establishing a strong foundation for extensions to domain-specific applications such as medical imaging, image super-resolution, and other specialized computer vision tasks. 

Our work opens promising directions for future research on alternative scanning and representation strategies for SSM in vision, which deviate from predefined directional adaptations in this field and can potentially lead to enhanced Mamba-based vision models.